\documentclass{article}
\usepackage{spconf,amsmath,graphicx}
\usepackage{amsmath}
\usepackage{graphicx}
\usepackage{url}
\usepackage{float}
\usepackage{hyperref}
\usepackage{stfloats}
\usepackage{caption}
\usepackage{svg}
\usepackage{amssymb}
\usepackage{placeins}
\usepackage{svg}
\usepackage{adjustbox}
\usepackage{tabularx}
\usepackage{caption}
\usepackage{array}
\usepackage{placeins} 

\usepackage{enumitem}
\setlist{nosep, leftmargin=14pt}

\usepackage{mwe} 

\title{Edge-boosted graph learning for Functional Brain Connectivity Analysis}
\name{David Yang$^{\star}$ \qquad Mostafa Abdelmegeed$^{\dagger}$ \qquad John Modl$^{\ddagger}$ \qquad Minjeong Kim$^{\dagger}$}
\address{$^{\star}$Department of Computer Science, Emory University, GA, USA \\
$^{\dagger}$Department of Computer Science, University of North Carolina at Greensboro, NC, USA\\$^{\ddagger}$Department of Computer Science and Engineering, University of Minnesota Twin Cities, MN, USA} 

\begin{document}
%
\maketitle
\begin{abstract}
Predicting disease states from functional brain connectivity is critical for the early diagnosis of severe neurodegenerative diseases such as Alzheimer's Disease and Parkinson's Disease. Existing studies commonly employ Graph Neural Networks (GNNs) to infer clinical diagnoses from node-based brain connectivity matrices generated through node-to-node similarities of regionally averaged fMRI signals. However, recent neuroscience studies found that such node-based connectivity does not accurately capture ``functional connections" within the brain. This paper proposes a novel approach to brain network analysis that emphasizes edge functional connectivity (eFC), shifting the focus to inter-edge relationships. Additionally, we introduce a co-embedding technique to integrate edge functional connections effectively. Experimental results on the ADNI and PPMI datasets demonstrate that our method significantly outperforms state-of-the-art GNN methods in classifying functional brain networks.
\end{abstract}

\section{Introduction}
\label{sec}
Neurodegenerative diseases, such as Alzheimer's disease (AD) and Parkinson's disease (PD), are among the most prevalent disorders affecting the brain and nervous system. Symptoms include memory loss, cognitive decline, behavioral changes, and motor dysfunction, ultimately leading to life-threatening conditions \cite{articl2e}. The prevalence of these diseases is rising, with nearly 7 million and 1  million Americans currently living with AD and PD, respectively, and these numbers are projected to increase significantly \cite{article}. As neither disease has a cure and current treatments merely slow progression, early detection at prodromal stages is essential.

Traditional diagnostic methods for neurodegenerative diseases rely heavily on clinical assessments and imaging data. However, recent advances in GNNs offer promising avenues for early detection by enabling the analysis of complex, non-Euclidean brain network data \cite{article7}. Studies have focused on the early stage of mild cognitive impairment (EMCI) of AD where an effective treatment could be applied to control or delay\cite{article3}. In neuroimaging studies, the human brain is often modeled as a network, with regions of interest (ROIs) serving as nodes and functional or structural connections between them as edges. These brain networks, derived from techniques such as functional magnetic resonance imaging (fMRI) or diffusion tensor imaging (DTI), reveal critical insights into the brain's organizational and connectivity patterns \cite{NENTWICH2020117001}.\\ 
\indent
Early detection of neurodegenerative diseases using graph learning typically begins with constructing a brain's functional connectivity matrix. Most methods employ node-based functional connectivity (nFC) \cite{LUO2022118792}, defined at the region level, to generate connectivity matrices. Here, brain regions are represented as nodes, with edges reflecting activity-based similarities, providing a static view of node-to-node interactions \cite{LI2021102233}. However, recent neuroscience research suggests that nFC may not fully represent connectivity, as it approximates region-to-region similarity rather than directly capturing connections. Neuroscientific evidence increasingly supports eFC, which directly represents dynamic relationships between edges. Studies have demonstrated that eFC matrices outperform traditional nFC matrices in tasks such as brain community detection \cite{faskowitz_edge-centric_2020}.\\
\indent
Most existing GNN methods are based on node-level connectivity matrices, typically calculated using Pearson's correlation coefficients to measure similarity between averaged signals at the region level—an indirect representation of connections between nodes. The recent neuroscientific definition of functional connections based on edge feature measurements. To bridge the gap between these GNN methods and the recent neuroscientific studies \cite{faskowitz_edge-centric_2020} suggesting edge focused calculations, we introduce a novel GNN architecture. Specifically, our GNN model co-embeds edge-to-edge relationships (edge attributes) directly from functional signals and node-to-node relationships (node attributes) from connectivity matrices in a unified framework as shown in Fig.1. This approach more accurately reflects the brain's interconnected nature, where the dynamic interplay between pathways is essential to understanding cognitive processes and behaviors. \\
\indent
We evaluate our model using data from the Alzheimer's Disease Neuroimaging Initiative (ADNI) and Parkinson's Progression Markers Initiative (PPMI) databases. By incorporating edge-to-edge relationships, our model enhances the functional representation capabilities of GNNs, aligning them more closely with the brain's intrinsic connectivity and improving the early detection of neurodegenerative diseases.

\section{Method}
\label{sec:typestyle}
\subsection{Edge Functional Connectivity}
We generate edge-attribute matrices from time series fMRI signals that record brain network activity across \(N\) regions, represented as nodes, over \(T\) seconds. The data matrix \(X\) contains time series \(x_n^{t}\) over the acquisition time \(t \in \{1, \ldots, T\}\) of temporal signals for each brain region \(n \in \{1, \ldots, N\}\).

Note that the traditional way of using the Pearson correlation coefficient averages temporal signals between regions, resulting in \(\hat{x}_n\) where \(n = 1, \ldots, N\). In contrast, we generate an edge matrix from the time series that keeps brain network activity over \(T\) across \(N\) regions. When we have time series at two different regions \(i, j \in N\), we multiply the \(z\)-scores \(z_i\) and \(z_j\) of these two rows \(x_i\) and \(x_j\), resulting in edge time signals (eTS) \(r_{ij} = z_i \cdot z_j\), which is a vector with the length \(T\). This calculation is done for all pairs of nodes (i.e., regions in the brain).

Fig. 1(a) shows an eTS matrix with the dimension of \(T \times N_e\), where $N_e$ is \(N \times (N-1)\) the total number of possible region combinations. Each element in this matrix represents the co-fluctuation magnitude between pairs of brain regions. The value of these co-fluctuations is positive when the activity of two regions is moving in the same direction at the same time. 

To incorporate the raw edge representations, eTS, into GNNs, we construct the eFC matrix. The eFC matrix captures the dependency of each edge, reflecting the similarity in their co-fluctuations. The definition of the eFC matrix $E_{FC}$is:

\begin{equation}
E_{FC} = \frac{E_{TS}^T \cdot E_{TS}}{\sqrt{d} \cdot \sqrt{d}^T},
\end{equation}

where $E_{TS}$ denotes edge time signals eTS and $d = \text{diag}(E_{TS}^T \cdot E_{TS}).$ Dot product is the normalization matrix obtained by the outer product of the standard deviation vector $c$ with itself, where the elements of \( d \cdot d^T \) range from \(\sigma_{\min}^2\) to \(\sigma_{\max}^2\). As shown in Fig. 1(b), this results in a matrix \(N_e \times N_e\), where each entry represents the dependency of one edge relative to all other edges.

Given its high dimensionality of the edge matrices and the inconsistency with node features, we further employ a co-embedding method to integrate the eFC matrix into a unified graph representation suitable for training GNNs as follows.

\begin{figure}[t]
    \centering
    \hspace{0.4cm} 
    \includegraphics[trim=0cm 0cm 0cm 0cm, clip, width=\columnwidth]{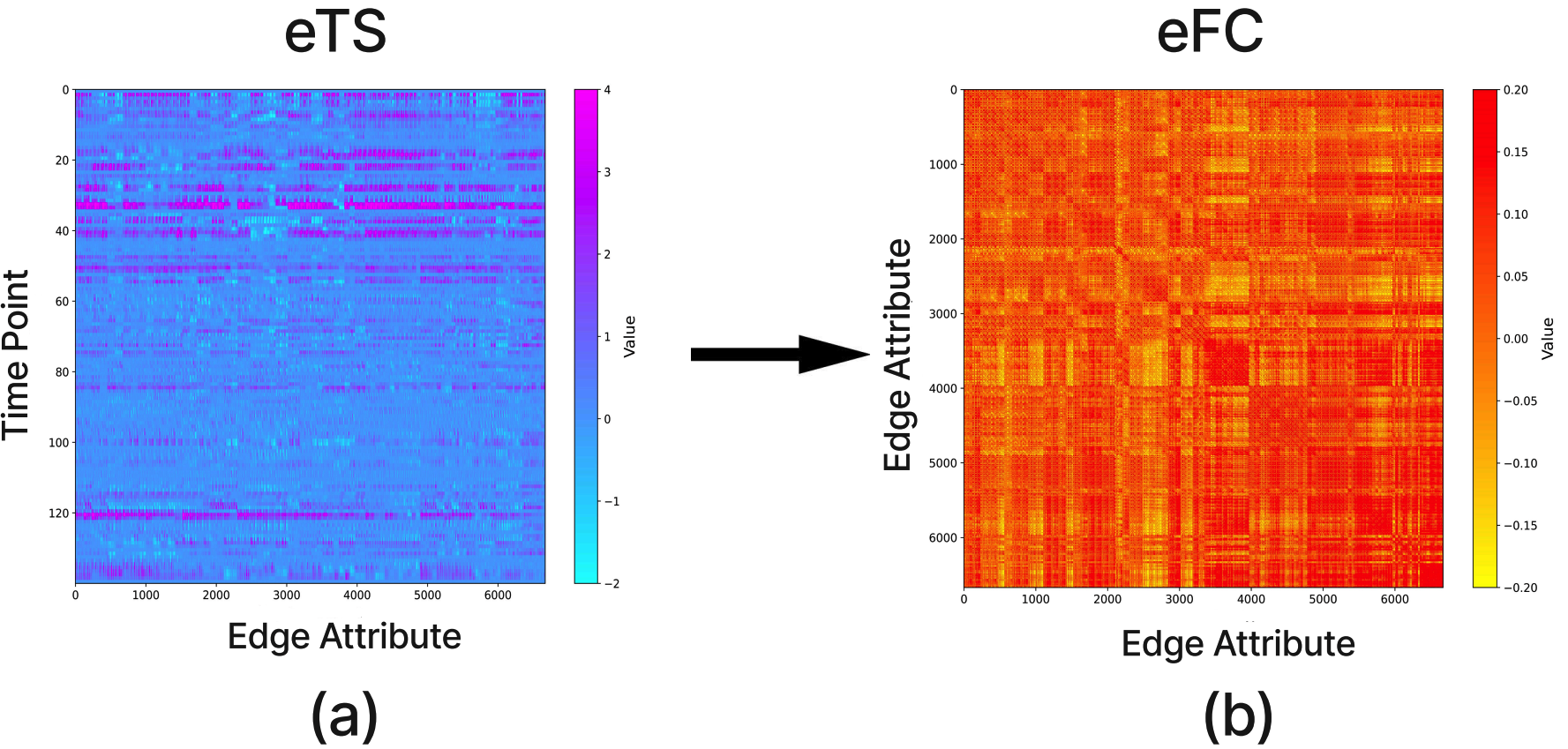}
    \vspace{-0.7cm} 
    \captionsetup{width=\columnwidth}
    \caption{Edge features in our method: (a) edge time series (eTS) with the size of $T \times N_e$, and (b) eFC matrix with the size of $N_e \times N_e$.}
    \label{fig:edge_features}
\end{figure}


\subsection{Co-embedding of Node and Edge Attributes}

Traditional GNN approaches often focus on node features only and are susceptible to the loss of information that is contained in edges, leading to sub-optimal representations. While node-based connectivity measures capture co-activation between regions through node-to-node feature similarity, edge-based connectivity reveals high-order relationships between regions. For instance, edge-based clustering allows a node to belong to multiple clusters, a flexibility not afforded by node-based clustering. Incorporating both node and edge features is essential, as they offer complementary perspectives on fMRI data.

We propose a co-embedding approach that integrates both nodes and edges as inputs, merging them into a unified node-edge representation. In GNN models, the hidden layers update the node feature representations by aggregating node features from the previous layer with the edge feature representations. Inspired by recent co-embedding approaches for GNNs \cite{zhou_co-embedding_2023}, we take nodes and edges as the input and combining them into a node-edge representation, updating the node embedding with the edge embedding layer. It is designed to be a unified framework, where layers compute low-dimensional representations for nodes by considering both the node's features and the features of their connecting edges.

Our model aggregates node representations and edge representations with the node features from the previous layer. Node-edge representations, which allow the model to better capture relationships between local nodes and edges, can lead to more accurate predictions.We represent a brain graph $G=(V,E)$ consisting of a finite set $V=\{v_i|i=1,...,N\}$ of $N$ nodes and edges $E \subset V \times V$, and a weighted adjacency matrix $W=[w_{i,j}]_{i,j=1}^N$ measuring the strength of each edge. $N = |V|$ and $N_e = |E|$ denote the number of nodes and edges, respectively. 

Let $A_v \in \mathbb{R}^{N \times N}$ be the node feature matrix derived from the correlation matrix of brain regions of the node, and $A_e \in \mathbb{R}^{N_e \times N_e}$ be the eFC matrix. $A_e(m, n)$ ranges from -1 to 1, where 1 indicates a positive connection, 0 indicates no connection, and -1 indicates a negative (opposite) connection, where $m, n \in \{1, \ldots, N_e\}$.



In Graph Convolutional Networks (GCNs) methods, each hidden layer aggregates features from neighboring nodes (defined by the adjacency matrix) and the outputs of the previous layer. Thus, the feature representation $H_i^{(l+1)}$ at the layer $l$+1 is defined as:

\begin{equation}
    H_i^{(l+1)} = \sigma \left( H^{(l)} \left( W_0^{(l)} + A \, W_1^{(l)} \right) \right), 
\end{equation}

where $A$ is the adjacency matrix.$W_0$ and $W_1$ are learnable weights.

However, in our co-embedding model, edge features are also integrated into the node embeddings. Our objective is to produce a matrix with the dimension of $V \times V$, which integrates both the node and the edge features. To address the challenges associated with the high dimensionality of edge features and inconsistencies between node and edge features, we propose direct co-embedding of the original edge matrices.

The feature representation $H_i^{(l+1)}$ of node $i$ in the $(l+1)$-th layer is updated by incorporating edge embeddings as follows:
\begin{equation}
H_i^{(l+1)} = \sigma \left( H^{(l)} \left[ W_0^{(l)} + \Phi(E_{efc}) \, W_1^{(l)} \right] \right).
\end{equation}

For each pair of connected nodes ${u,v} \in E_{efc}$, $E_{efc}^{u,v} \in \mathbb{R}$ represent the feature vector of the edge between nodes $u$ and $v$. Here, $\Phi$ is a function that retrieves edge features associated with the node representation from the edge functional connectivity $E_{efc}$.

\section{Experimental results}
\label{sec:majhead}

\subsection{Data}

We evaluated our method on fMRI data in the ADNI\footnote{https://adni.loni.usc.edu/} and the PPMI\footnote{https://www.ppmi-info.org/} databases.

The ADNI dataset consists of fMRI scans of 250 patients, including 27 AD patients, 75 CN patients, 69 early MCI patients, 46 late MCI patients and 33 SMC patients. CN, early MCI, and SMC patients are considered patients without the disease, while AD and late MCI patients are classified as patients with the disease.
The PPMI dataset consists of fMRI scans of 209 patients, including 15 control and 114 in the experimental group, with 67 prodromal and 14 SWEDD. Each patient is classified as a separate label during the preprocess state. All fMRI data was processed by using fMRIPrep \cite{buckner2013opportunities} for preprocessing and BOLD signal extraction.

\subsection{Classification Performance}

In our method, we used two layers of modified graph convolutional layers 
\cite{9224195}, where one is the node layer aggregating edge information to node representation, and the other one is the edge layer aggregating node information to edge representation. The experiment was done with Epoch 300, with a learning rate of 0.0001, weight decay of 0.0005, a dropout rate of 0.5, and a hidden dimension of 1024. 

We compared the classification performance on both datasets using our method with the results of CNN, GCN \cite{kipf2017semisupervised}, CRGNN \cite{inbook}, and MGNN \cite{kanatsoulis2024counting}.

\begin{itemize}

\item {CNN: We used two layers of 1D convolution with batch normalization. The experiment was conducted over 300 epochs with a learning rate of 0.0001, weight decay of 0.0005, dropout rate of 0.5, and a hidden dimension of 512.}

\item {GCN: The experiment used two graph convolutional layers as described in \cite{kipf2017semisupervised}. The experiment was conducted over 300 epochs with a learning rate of 0.0001, a weight decay of 0.0005, a dropout rate of 0.3, and a hidden dimension of 512.}

\item {CRGNN \cite{inbook}: The experiment had 300 epochs, $\lambda_1$ and $\lambda_1$ as 0.0001, 116 nodes, respectively, and a learning rate of 0.001.}

\item {MGNN \cite{kanatsoulis2024counting}: The experiment was conducted over 300 epochs, with a learning rate of 0.0001, weight decay of 0.0005, a dropout rate of 0.5, and a hidden dimension of 512. For topological features, we used $dim$=45 and $K$=10 (i.e., the features are calculated up to the 10th order).}

\end{itemize}
\FloatBarrier

\begin{table}[H]
\centering
\captionsetup{labelformat=empty}
\caption{\textbf{Table 1.} 10-fold cross-validation results on ADNI dataset to compare classification performances.}
\resizebox{\columnwidth}{!}{
\begin{tabular}{|c|c|c|c|}
\hline
\textbf{Model} & \textbf{Accuracy} & \textbf{Precision} & \textbf{F1 Score} \\ \hline
\textbf{CNN} & $0.7640 \pm 0.0413$ & $0.7510 \pm 0.1167$ & $0.7330 \pm 0.0711$ \\ \hline
\textbf{GCN} & $0.7440 \pm 0.0413$ & $0.7420 \pm 0.1167$ & $0.7194 \pm 0.0711$ \\ \hline
\textbf{CRGNN} & $0.7680 \pm 0.0620$ & $0.7665 \pm 0.1370$ & $0.6196 \pm 0.0808$ \\ \hline
\textbf{MGNN} & $0.7680 \pm 0.0534$ & $0.7680 \pm 0.0561$ & $0.7249 \pm 0.0693$ \\ \hline
\textbf{Our method} & \underline{$0.8000 \pm 0.0876$} & \underline{$0.8437 \pm 0.0614$} & \underline{$0.7659 \pm 0.1181$} \\
\hline
\end{tabular}}
\end{table}
\begin{table}[H]
\centering
\captionsetup{labelformat=empty}
\caption{\textbf{Table 2.} 10-fold cross-validation results on PPMI dataset to compare classification performances.}
\resizebox{\columnwidth}{!}{
\begin{tabular}{|c|c|c|c|}
\hline
\textbf{Model} & \textbf{Accuracy} & \textbf{Precision} & \textbf{F1 Score} \\ \hline
\textbf{CNN} & $0.6802 \pm 0.1075$ & $0.6533 \pm 0.1350$ & $0.6141 \pm 0.1342$ \\ \hline
\textbf{GCN} & $0.6802 \pm 0.1157$ & $0.6028 \pm 0.1809$ & $0.6156 \pm 0.1525$ \\ \hline
\textbf{CRGNN} & $0.6556 \pm 0.0808$ & $0.5882 \pm 0.0924$ & $0.5111 \pm 0.0882$ \\ \hline
\textbf{MGNN} & $0.6945 \pm 0.1037$ & $0.6310 \pm 0.1532$ & $0.6323 \pm 0.1329$ \\ \hline
\textbf{Our method} & \underline{$0.7083 \pm 0.0572$} & \underline{$0.6821 \pm 0.0860$} & \underline{$0.6700 \pm 0.0587$} \\
\hline
\end{tabular}}
\end{table}

\FloatBarrier 

Table 1 and 2 present the classification performances of our method and four comparison methods on the ADNI and PPMI datasets, respectively. Our method consistently outperforms all four methods on both datasets across three different metrics, including Accuracy, Precision, and F1 Score, with particularly substantial gains observed on the ADNI dataset. The superior performance of our model is attributed to its unique capability to jointly embed edge and node attributes, effectively preserving the rich structural and functional characteristics of brain connectivity patterns that are often overlooked by traditional approaches.

\section{Conclusion}
Extracting the complex dynamics of disease progression to predict disease states from brain functional networks is a significant challenge. To address this, we introduced a novel GNN architecture that captures brain functional connectivity patterns associated with neurodegeneration by incorporating new edge-based functional connection information alongside node feature representations. Experiments on two benchmark datasets—ADNI and PPMI—demonstrate the effectiveness of our approach in disease classification tasks, outperforming widely used GNN methods. These results validate our method's utility and highlight promising future research directions in neuroscience and clinical applications. Our work contributes to the growing evidence that advanced neural network architectures can markedly enhance our ability to diagnose and understand neurodegenerative disorders when designed to integrate domain-specific knowledge.

\section{Acknowledgments}
\label{sec:acknowledgments}
This work was done during the REU (Research Experiences for Undergraduates) program in 2024 at UNC Greensboro, supported by NSF Grant CNS-2349369.

\bibliographystyle{IEEEbib}
\bibliography{strings}

\begin{thebibliography}{10}

\bibitem{articl2e}
Connie Marras, Eric Roberts, Caroline Tanner, Stephen Van~Den Eeden, Rodolfo Savica, Web Ross, Brian Fiske, James Beck, and Allison~Wright Willis,
\newblock ``Incidence of parkinson disease in north america (p1-1.virtual),''
\newblock {\em Neurology}, vol. 98, no. 18\_supplement, pp. 2665, 2022.

\bibitem{article}
Zhao Longhe,
\newblock ``2020 alzheimer’s disease facts and figures,''
\newblock {\em Alzheimer's \& Dementia}, vol. 16, 06 2020.

\bibitem{article7}
Sheng Liu, Arjun Masurkar, Henry Rusinek, Jingyun Chen, Ben Zhang, Weicheng Zhu, Carlos Fernandez-Granda, and Narges Razavian,
\newblock ``Generalizable deep learning model for early alzheimer’s disease detection from structural mris,''
\newblock {\em Scientific Reports}, vol. 12, 2022.

\bibitem{article3}
Janani Venugopalan, Li~Tong, Hamid~Reza Hassanzadeh, and May Wang,
\newblock ``Multimodal deep learning models for early detection of alzheimer’s disease stage,''
\newblock {\em Scientific Reports}, vol. 11, pp. 3254, 02 2021.

\bibitem{NENTWICH2020117001}
Maximilian Nentwich, Lei Ai, Jens Madsen, Qawi~K. Telesford, Stefan Haufe, Michael~P. Milham, and Lucas~C. Parra,
\newblock ``Functional connectivity of eeg is subject-specific, associated with phenotype, and different from fmri,''
\newblock {\em NeuroImage}, vol. 218, pp. 117001, 2020.

\bibitem{LUO2022118792}
Wenjing Luo and R.~Todd Constable,
\newblock ``Inside information: Systematic within-node functional connectivity changes observed across tasks or groups,''
\newblock {\em NeuroImage}, vol. 247, pp. 118792, 2022.

\bibitem{LI2021102233}
Xiaoxiao Li, Yuan Zhou, Nicha Dvornek, Muhan Zhang, Siyuan Gao, Juntang Zhuang, Dustin Scheinost, Lawrence~H. Staib, Pamela Ventola, and James~S. Duncan,
\newblock ``Braingnn: Interpretable brain graph neural network for fmri analysis,''
\newblock {\em Medical Image Analysis}, vol. 74, pp. 102233, 2021.

\bibitem{faskowitz_edge-centric_2020}
Joshua Faskowitz, Farnaz~Zamani Esfahlani, Youngheun Jo, Olaf Sporns, and Richard~F. Betzel,
\newblock ``Edge-centric functional network representations of human cerebral cortex reveal overlapping system-level architecture,''
\newblock {\em Nature Neuroscience}, vol. 23, no. 12, pp. 1644--1654, Dec. 2020.

\bibitem{zhou_co-embedding_2023}
Yuchen Zhou, Hongtao Huo, Zhiwen Hou, Lingbin Bu, Jingyi Mao, Yifan Wang, Xiaojun Lv, and Fanliang Bu,
\newblock ``Co-embedding of edges and nodes with deep graph convolutional neural networks,''
\newblock {\em Scientific Reports}, vol. 13, no. 1, pp. 16966, Oct. 2023.

\bibitem{buckner2013opportunities}
Randy~L Buckner, Fenna~M Krienen, and BT~Thomas Yeo,
\newblock ``Opportunities and limitations of intrinsic functional connectivity mri,''
\newblock {\em Nature neuroscience}, vol. 16, no. 7, pp. 832--837, 2013.

\bibitem{9224195}
Xiaodong Jiang, Ronghang Zhu, Pengsheng Ji, and Sheng Li,
\newblock ``Co-embedding of nodes and edges with graph neural networks,''
\newblock {\em IEEE Transactions on Pattern Analysis and Machine Intelligence}, vol. 45, no. 6, pp. 7075--7086, 2023.

\bibitem{kipf2017semisupervised}
Thomas~N. Kipf and Max Welling,
\newblock ``Semi-supervised classification with graph convolutional networks,''
\newblock in {\em International Conference on Learning Representations}, 2017.

\bibitem{inbook}
Xia Zhengwang, Huan Wang, Tao Zhou, Zhuqing Jiao, and Jianfeng Lu,
\newblock {\em Customized Relationship Graph Neural Network for Brain Disorder Identification}, pp. 109--118,
\newblock 2024.

\bibitem{kanatsoulis2024counting}
Charilaos Kanatsoulis and Alejandro Ribeiro,
\newblock ``Counting graph substructures with graph neural networks,''
\newblock in {\em The Twelfth International Conference on Learning Representations}, 2024.

\end{thebibliography}

\end{document}